\documentclass[nohyperref]{article}

\usepackage{microtype}
\usepackage{graphicx}
\usepackage{subfigure}
\usepackage{booktabs} % for professional tables

\usepackage[utf8]{inputenc} % allow utf-8 input
\usepackage[T1]{fontenc}    % use 8-bit T1 fonts
\usepackage[colorlinks]{hyperref}          % hyperlinks
\usepackage{url}            % simple URL typesetting
\usepackage{booktabs}       % professional-quality tables
\usepackage{amsfonts}       % blackboard math symbols
\usepackage{nicefrac}       % compact symbols for 1/2, etc.
\usepackage{microtype}      % microtypography
\usepackage{xcolor}         % colors
\usepackage{bbm, dsfont}
\usepackage{wrapfig}
\usepackage{wrapfig}
\usepackage{graphicx}
\usepackage{multirow}
\usepackage{booktabs}
\usepackage{tikz}
\usepackage{comment}
\usepackage{amsmath,amssymb} % define this before the line numbering.
\usepackage{color}
\usepackage{bm}
\usepackage{makecell}
\usepackage{colortbl}
\definecolor{gray}{gray}{.15}
\usepackage{caption}
\usepackage{subcaption}  % This package is incompatible with the subfigure and subfig packages.
\usepackage{enumerate}
\usepackage{framed}

\include{command}
\usepackage{mathabx}    % math
\usepackage{algorithm,algpseudocode}

\usepackage[export]{adjustbox}
\newcommand{\highlight}[1]{\cellcolor{gray!16}\textbf{#1}}
\usepackage{hyperref}

\usepackage[accepted]{icml2024}

\usepackage{amsmath}
\usepackage{amssymb}
\usepackage{mathtools}
\usepackage{amsthm}

\usepackage[capitalize,noabbrev]{cleveref}

\theoremstyle{plain}

\theoremstyle{definition}

\theoremstyle{remark}

\usepackage[textsize=tiny]{todonotes}

\icmltitlerunning{Debiasing Text-to-Image Diffusion Models}

\definecolor{mydarkblue}{rgb}{0,0.1,0.6}
\hypersetup{
    colorlinks=true,
    citecolor=mydarkblue
          }

\begin{document}

\twocolumn[
\icmltitle{Debiasing Text-to-Image Diffusion Models}



\icmlsetsymbol{equal}{*}

\author{
Ruifei He\textsuperscript{\rm 1*} 
\;\; Shuyang Sun\textsuperscript{\rm 2} 
\;\; Xin Yu\textsuperscript{\rm 1} 
\;\; Chuhui Xue\textsuperscript{\rm 3} 
\;\; Wenqing Zhang\textsuperscript{\rm 3} 
\;\; Philip Torr\textsuperscript{\rm 2} \\
\;\textbf{Song Bai}\textsuperscript{\rm 3$\dagger$}
\;\; \textbf{Xiaojuan Qi}\textsuperscript{\rm 1$\dagger$}
\\
\;\textsuperscript{\rm 1}The University of Hong Kong \ \textsuperscript{\rm 2}University of Oxford \ \textsuperscript{\rm 3}ByteDance 
}

\begin{icmlauthorlist}
\icmlauthor{Ruifei He}{1}\textsuperscript{\rm*} 
\icmlauthor{Chuhui Xue}{2}
\icmlauthor{Haoru Tan}{1}
\icmlauthor{Wenqing Zhang}{2}
\icmlauthor{Yingchen Yu}{2}
\icmlauthor{Song Bai}{2}\textsuperscript{\rm$\dagger$}
\icmlauthor{Xiaojuan Qi}{1}\textsuperscript{\rm$\dagger$}

\textsuperscript{\rm 1}The University of Hong Kong \ \textsuperscript{\rm 2}ByteDance 
\end{icmlauthorlist}



\icmlkeywords{Machine Learning, ICML}

\vskip 0.3in
]
\begin{figure*} [t!]
% \vspace{-0.95cm}
   \begin{center}
   % \fbox{\rule{0pt}{2in} \rule{.9\linewidth}{0pt}} 
   \includegraphics[width=1.0\linewidth]{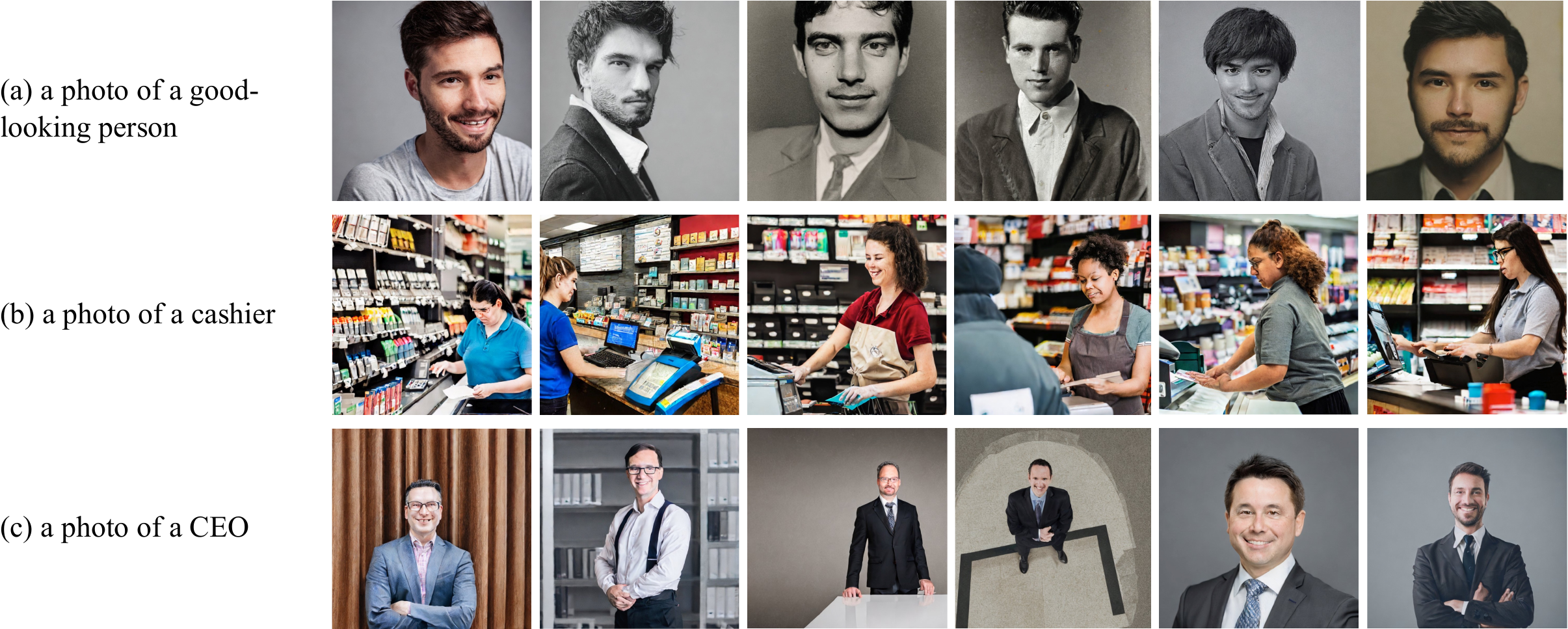} 
   \end{center}
    % \vspace{-0.4cm}
      \caption{Examples of \textbf{social bias in generated images} by text-to-image diffusion models. Left is the input prompt. (a) The prompt "Good-looking person" leads to mostly white young males; (b) The prompt of a low-paying occupation "cashier" generates images of all females; (c) The prompt of a high-paying occupation "CEO" generates images of biased towards whiteness and masculinity.}
   \label{fig: bias eg} 
   % \vspace{-0.5cm} 
   \end{figure*}
   


% \printAffiliationsAndNotice{}

{\let\thefootnote\relax\footnotetext{$^*$ Part of the work is done during an internship at ByteDance. Email: \href{mailto:ruifeihe@eee.hku.hk}{\color{black}{ruifeihe@eee.hku.hk}}}}
{\let\thefootnote\relax\footnotetext{$^\dagger$ Corresponding authors: 
\href{mailto:songbai.site@gmail.com}{\color{black}{songbai.site@gmail.com}}, \href{mailto:xjqi@eee.hku.hk}{\color{black}{xjqi@eee.hku.hk}}}}
{\let\thefootnote\relax\footnotetext{Preprint. Under review. }}

\begin{abstract}
Learning-based Text-to-Image (TTI) models like Stable Diffusion have revolutionized the way visual content is generated in various domains. However, recent research has shown that nonnegligible social bias exists in current state-of-the-art TTI systems, which raises important concerns. In this work, we target resolving the social bias in TTI diffusion models. We begin by formalizing the problem setting and 
%in light of Safe Latent Diffusion \citep{schramowski2023safe}, we 
use the text descriptions of bias groups to establish an unsafe direction for guiding the diffusion process. Next, we simplify the problem into a weight optimization problem and
attempt a Reinforcement solver, Policy Gradient, which shows sub-optimal performance with slow convergence. Further, to overcome limitations, we propose an iterative distribution alignment (IDA) method. Despite its simplicity, we show that IDA shows efficiency and fast convergence in resolving the social bias in TTI diffusion models. Our code will be released.

%We begin by formalizing the problem and leveraging Safe Latent Diffusion (Schramowski, 2023) to establish an unsafe direction for guiding the diffusion process using text descriptions of biased groups. Subsequently, we simplify the problem into a weight optimization problem and explore the use of a Reinforcement solver, Policy Gradient, which exhibits sub-optimal performance with slow convergence. To overcome these limitations, we propose an iterative distribution alignment (IDA) method. Through extensive experimentation, we demonstrate that IDA efficiently resolves social bias in TTI diffusion models with fast convergence. The code for our approach will be made publicly available.
\end{abstract}

\section{Introduction}
State-of-the-art Text-to-Image models such as Stable Diffusion \citep{rombach2022high},  DALL-E2 \citep{ramesh2022hierarchical}, Imagen \citep{saharia2022photorealistic} and GLIDE \citep{nichol2021glide} have become a trend in prompted image generation. These models take a natural language as an input prompt, and output images consistent with the prompt. 

However, recent work \cite{luccioni2023stable} points out that these generated data could exhibit strong social bias, which could originate from different sources: 1) the training data scraped from the Web could be biased; 2) the bias inherited from the CLIP model \cite{radford2021learning} since many text-to-image models use it to guide the generation process.

To help the audience better understand the bias, we ask Stable Diffusion \citep{rombach2022high} to generate some illustrations for us. In Figure \ref{fig: bias eg} (a), the prompt is ``a photo of a good-looking person'', and the generated images are significantly biased towards whiteness and masculinity. In Figure \ref{fig: bias eg} (b) and (c), we ask the model to generate images of low-paying (cashier) and high-paying (CEO) occupations, where we can see the model has a stereotype that high-paying occupations are mostly white males while low-paying ones are more often females. We emphasize this to be a serious concern from the ethical perspective.

\begin{figure*} 
% \vspace{-0.95cm}
   \begin{center}
   % \fbox{\rule{0pt}{2in} \rule{.9\linewidth}{0pt}} 
   \includegraphics[width=1.0\linewidth]{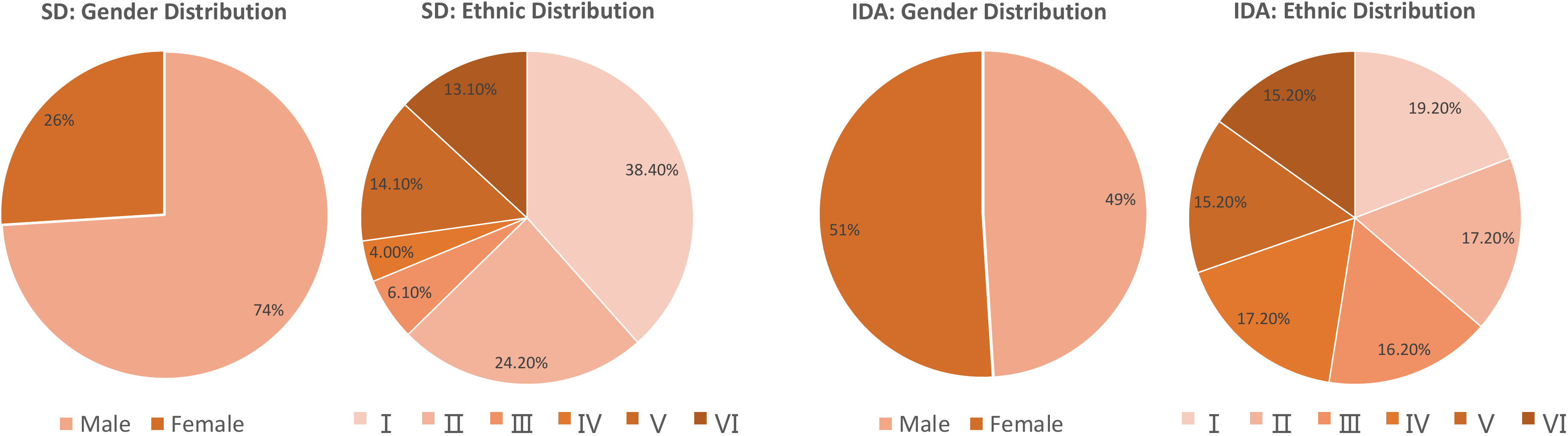} 
   \end{center}
    % \vspace{-0.4cm}
      \caption{\textbf{Left: Social Bias in Stable Diffusion (SD)}: gender and ethnic bias. For ethnicity, the group \uppercase\expandafter{\romannumeral1} to \uppercase\expandafter{\romannumeral6} represents skin color from light to dark. The gender distribution is largely biased towards males, and the ethnic distribution is also significantly biased with whites being the majority. \textbf{Right: Debiasing Social Bias by IDA}: with the proposed iterative distribution alignment (IDA) method, both gender and ethnic distribution are redistributed to a balanced distribution.}
   \label{fig: bias} 
   % \vspace{-0.5cm} 
   \end{figure*}

Except from visual examples, we also conduct statistical experiments with Stable Diffusion and find it exists strong social bias on both gender and ethnicity. 
As shown in Figure \ref{fig: bias} Left, for gender, it has a preference of 74\% for males than females; for ethnicity, we adopt the Fitzpatrick Skin Scale classification system \citep{fitzpatrick1988validity} (divides people into six groups by skin type) and find the model strongly biased towards whiteness. 
As more and more users are using these models, millions of biased synthetic data are produced from these models, which we emphasize to be a serious concern.
Nevertheless, how to resolve these biases remains under-explored.

In this work, we target resolving the social bias in text-to-image diffusion models. We first formalize the problem setting, and in light of Safe Latent Diffusion \citep{schramowski2023safe}, we use the text descriptions of biased groups to form an unsafe direction to guide the diffusion process. Since there could be more than two groups, we turn the problem into a weight optimization problem where the weights are the coefficients of text descriptions for each biased group.
Due to the nature of the problem being a black-box optimization task, we first resort to a reinforcement framework, Policy Gradient \citep{sutton1999policy}. Though it manages to reduce the bias, its slow convergence and sub-optimal performance make it impractical to use in realistic situations.  
Further, we propose an iterative distribution alignment (IDA) method to match the biased distribution to a unique distribution. 

We conduct experiments on the Stable Diffusion model and study how to resolve gender and ethnic bias. As shown in Figure \ref{fig: bias} Right, the proposed IDA method successfully redistributes the gender and ethnic distribution to a balanced distribution, achieving promising debiasing results.
Further, we explore gender bias in various occupations. With extensive experimental results, we show that IDA embraces efficiency and fast convergence in resolving the social bias in TTI diffusion models. 

\section{Related Works}
\paragraph{Text-to-image Diffusion models.}
Diffusion models \citep{sohl2015deep, ho2020denoising, nichol2021improved} have recently emerged as a class of promising and powerful generative models. 
As a likelihood-based model, the diffusion model matches the underlying data distribution $q(x_0)$ by learning to reverse a noising process, and thus novel images can be sampled from a prior Gaussian distribution via the learned reverse path. 
In particular, text-to-image generation can be treated as a conditional image generation task that requires the sampled image to match the given natural language description. Based upon the formulation of the diffusion model, several text-to-image models such as Stable diffusion \citep{rombach2022high}, DALL-E2 \citep{ramesh2022hierarchical}, Imagen \citep{saharia2022photorealistic} and GLIDE \citep{nichol2021glide} deliver unprecedented synthesis quality, largely facilitating the development of the AI-for-Art community.

\paragraph{Bias in Text-to-image Diffusion models.} Recently, a few studies explored social bias in text-to-image diffusion models. \citet{luccioni2023stable} explore and quantify the social biases in text-to-image systems by creating collections of generated images that highlight the variation of a text-to-image system across gender, ethnicity, professions, and gender-coded adjectives. Their results suggest that different TTI systems exhibit social bias towards whiteness and masculinity. \citet{naik2023social} take a multi-dimensional approach to studying and quantifying common social biases in the generated images across representations
of attributes such as gender, age, race, and geographical location, and find that significant biases exist across all attributes. To the best of our knowledge, no research has formally addressed the problem of resolving social bias in TTI diffusion models.

\section{Method}
\subsection{Background on Diffusion models} 
%感觉如果涉及到的符号如果很多，并且文章长度不够的话，可以加一个表格来叙述一下符号声明。
%感觉这段可能需要稍微改一下叙述方式让整体变的更加简洁易读。
%其实读者在这里可能更在乎的点是：
%1. SD是个什么，有哪些参数，参数有什么性质。SD其实是: 
%The Stable Diffusion (SD) model employs a noise-injection process, also known as the forward process, to generate data from noise. This process involves gradually introducing noise into the original clean data (represented as x0) and then reversing this process to generate data from the injected noise. 我觉得可以参考一下这篇文章的Diffusion的简介。https://openreview.net/pdf?id=7hxoYxKDTV
%2. 紧接着用一段叙述：SD的前向计算
%3. 紧接着用一段叙述：SD的去燥计算
%我觉得这个Lower-bound可以不用叙述。读者更关心的是详细的计算过程吧。
%
% \thr{Ruifei, please see the comments here.} 
The denoising diffusion probabilistic model (DDPM) employs a noise-injection process, also known as the forward process, to generate data from noise. This process involves gradually introducing noise into the original clean data (represented as $x_0$) and then reversing this process to generate data from the injected noise.

Formally, given a sample from the data distribution $x_0 \sim q(\mathbf{x}_0)$, a forward process $q\left(\mathbf{x}_{1: T} \mid \mathbf{x}_0\right)=\prod_{t=1}^T q\left(\mathbf{x}_t \mid \mathbf{x}_{t-1}\right)$ progressively perturbs the data with Gaussian kernels $q\left(\mathbf{x}_t \mid \mathbf{x}_{t-1}\right):=\mathcal{N}\left(\sqrt{1-\beta_t} \mathbf{x}_{t-1}, \beta_t \mathbf{I}\right)$, producing increasingly noisy latent variables $\mathbf{x}_{1}, \mathbf{x}_{2}, ..., \mathbf{x}_{T}$. Notably, $x_t$ can be directly sampled from $x_0$ thanks to the closed form:
\begin{equation}
q\left(\mathbf{x}_t \mid \mathbf{x}_0\right)=\mathcal{N}\left(\mathbf{x}_t ; \sqrt{\bar{\alpha}_t} \mathbf{x}_0,\left(1-\bar{\alpha}_t\right) \mathbf{I}\right)
\end{equation}
% where $\alpha_t:=1-\beta_t$ and $\bar{\alpha}_t:=\prod_{s=1}^t \alpha_s$. \thr{Maybe we should introduce these factors with more details? like the xxx is the xxx in the forward/inverse process and xxx is the original and xxx is the estimated one.} 
% In general, the forward process variances $\beta_t$ are fixed and increased linearly from $\beta_1=10^{-4}$ to $\beta_T=0.02$. Besides, $T$ should be large (e.g., 1000) enough to ensure $q\left(\mathbf{x}_T \mid \mathbf{x}_0\right)\approx \mathcal{N}(0, \mathbf{I})$. 

% Diffusion model aims to model the joint distribution $q\left(\mathbf{x}_{0:T}\right)$ which naturally involves a tractable sampling path for the marginal distribution $q\left(\mathbf{x}_{0}\right)$.

% Then, training is achieved by minimizing the variational upper bound of negative log-likelihood: 
% \begin{equation}
%    \mathrm{E}_{q\left(\mathbf{x}_0\right)}\left[-\log p_\theta\left(\mathbf{x}_0\right)\right] \leq \mathrm{E}_{q\left(\mathbf{x}_{0: T}\right)}\left[-\log \frac{p_\theta\left(\mathbf{x}_{0: T}\right)}{q\left(\mathbf{x}_{1: T} \mid \mathbf{x}_0\right)}\right]
% \end{equation}

For the reverse process, started from an initial noise map $x_T\sim p(\mathbf{x}_T)=\mathcal{N}(0, \mathbf{I})$, new images can be then generated via iteratively sampling from $p_\theta\left(\mathbf{x}_{t-1} \mid \mathbf{x}_t\right)$ using the following equation:
\begin{equation}
\mathbf{x}_{t-1}=\frac{1}{\sqrt{\alpha_t}}\left(\mathbf{x}_t-\frac{\beta_t}{\sqrt{1-\bar{\alpha}_t}} \boldsymbol{\epsilon}_\theta\left(\mathbf{x}_t, t\right)\right)+\sigma_t \mathbf{z}, 
\end{equation} where $\mathbf{z} \sim \mathcal{N}(\mathbf{0}, \mathbf{I}).$

The text-to-image diffusion model extends the basic unconditional diffusion model by changing the target distribution $q(\mathbf{x_0})$ into a conditional one $q(\mathbf{x_0}\mid \mathbf{c})$, where $\mathbf{c}$ is a natural language description.
Besides, following the improved DDPM (\cite{nichol2021improved}), $\boldsymbol{\Sigma}_\theta$ is also estimated in GLIDE (\cite{nichol2021glide}).

% Classifier guidance mainly relies on an extra-trained noise CLIP model to provide feedback at intermediate sampling steps. Classifier-free guidance, on the other hand, randomly drops the text prompt with a fixed probability $p$ during the training, which can be viewed as joint training of an unconditional model $\boldsymbol{\epsilon}_\theta\left(\mathbf{x}_t \mid \emptyset\right)$ (\textit{i.e.}, $\boldsymbol{\epsilon}_\theta\left(\mathbf{x}_t\right)$) and a conditional model $\boldsymbol{\epsilon}_\theta\left(\mathbf{x}_t \mid \mathbf{c}\right)$. At each sampling step, the model's output is actually performed using an extrapolation as follows:
% \begin{equation}
% \boldsymbol{\hat{\epsilon}}_\theta\left(\mathbf{x}_t \mid \mathbf{c}\right)=\boldsymbol{\epsilon}_\theta\left(\mathbf{x}_t \mid \emptyset\right)+s \cdot\left(\boldsymbol{\epsilon}_\theta\left(\mathbf{x}_t \mid \mathbf{c}\right)-\boldsymbol{\epsilon}_\theta\left(\mathbf{x}_t \mid \emptyset\right)\right)
% \end{equation}
% where $s$ is a guidance scale that can trade off sampling quality and diversity. 

\subsection{Problem Formulation}

\begin{figure*} 
% \vspace{-0.95cm}
   \begin{center}
   % \fbox{\rule{0pt}{2in} \rule{.9\linewidth}{0pt}} 
   \includegraphics[width=0.9\linewidth]{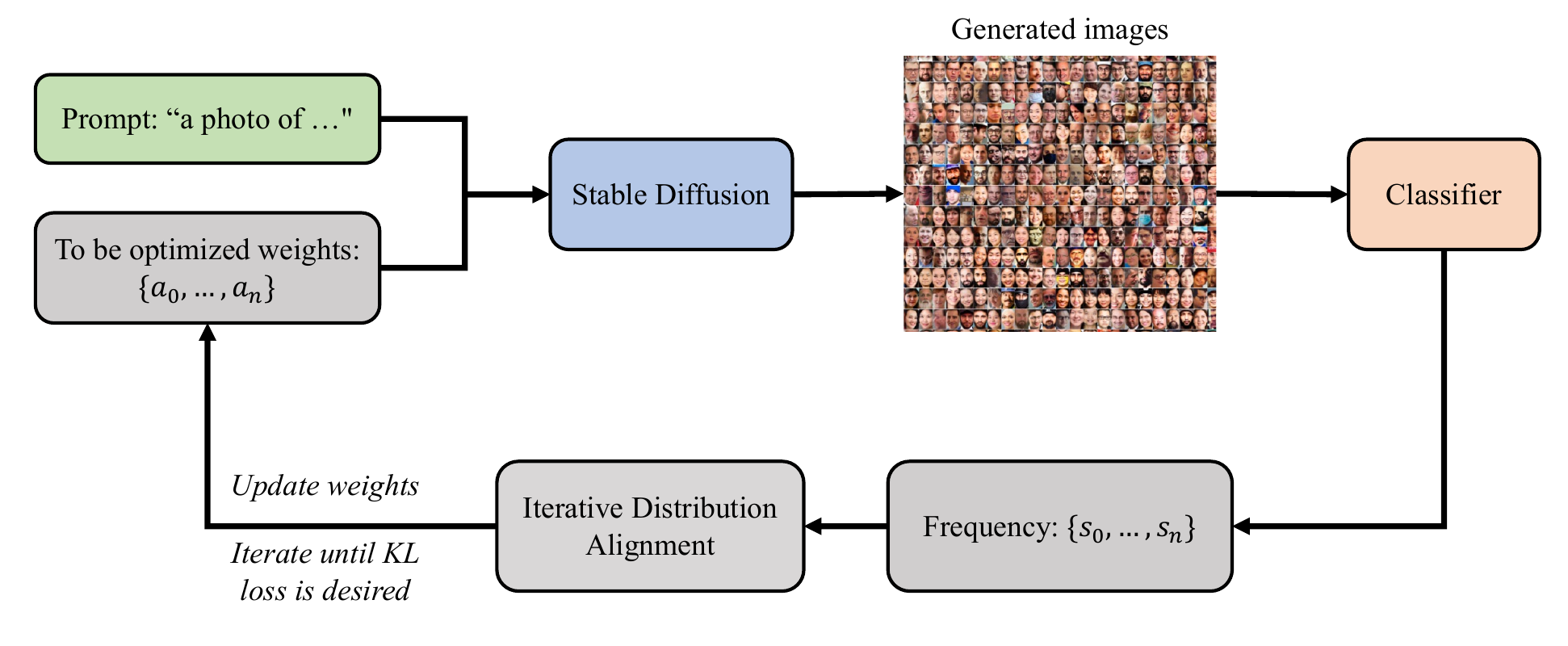} 
   \end{center}
    % \vspace{-0.4cm}
      \caption{\textbf{IDA framework}. We input the to-be-optimized weights and the prompt into Stable Diffusion to generate images, and use an automatic classifier to obtain the frequency. Next, we use IDA to update the weights by the frequency. We iterate this process until the KL divergence loss between the obtained frequency and a uniform distribution is below the desired value.}
   \label{fig: framework} 
   \vspace{-0.2cm} 
   \end{figure*}

% 可能还是有些地方我没有读太懂，可能是因为我本身对于扩散模型的理解还不够，我感觉可能可以考虑换一些叙述方式
%这里可能存在一个更好突出本文优势的叙述逻辑：
%1. 首先定义multi-directional guidance \bm{u} 和其中的weights。并介绍如何加入到公式6的。
%2. \bm{u}是如何参与到reverse计算过程的，可以把它添加到当前的公式5和6中，但是当前的公式5和6其实是没有找到u是如何参与其中的。
%3. 紧接着叙述加解释\bm{u}作为一个latent guidence是如何可以改变输出的。
%4. 讨论这一套pipelien与当前的现有工作，SLD的联系。

% \thr{Ruifei, please see the comments here.} 
% safe stable diffusion; our problem definition
We consider a diffusion model $SD$ that generates images from an input noise latent vector $\bm{z}$, a text prompt $\bm{p}$, and we hope to resolve its social bias with respect to a certain bias evaluation perspective, i.e. gender or ethnicity. We have $n$ evaluation attributes, with their text description encoded by a text encoder given by 
$\{\bm{g}_0, ..., \bm{g}_{n-1}\}$. 
Given a text prompt $\bm{p}$, 
we hope to produce an unbiased TTI system that could generate images with a uniform distribution across the given evaluation attributes. 

\paragraph{Safe Latent Diffusion} SLD \cite{schramowski2023safe} is an approach designed to reduce the inappropriate generated contents of diffusion models. To influence the diffusion process, SLD defines an inappropriate concept in addition to the text prompt and moves the unconditioned score estimate towards the prompt conditioned estimate and simultaneously away from the inappropriate concept conditioned estimate. 
Concretely, different from the usual Stable Diffusion's inference as: 
\begin{equation}\label{eq:classifier_free}
    \mathbf{\Tilde{\epsilon}}_\theta(\mathbf{z}_t, \mathbf{c}_p) := \mathbf{\epsilon}_\theta(\mathbf{z}_t) + s_g (\mathbf{\epsilon}_\theta(\mathbf{z}_t, \mathbf{c}_p) - \mathbf{\epsilon}_\theta(\mathbf{z}_t)), 
\end{equation}
where $\mathbf{c}_p$ is the text condition.  
It adds an inappropriate concept via textual description $S$ and obtains another conditioned estimate  $\mathbf{\epsilon}_\theta(\mathbf{z}_t, \mathbf{c}_S)$.
Hence, the adjusted inference estimates become
$ \mathbf{\Bar{\epsilon}}_\theta(\mathbf{z}_t, \mathbf{c}_p, \mathbf{c}_S)=$
\begin{align}
\label{eq:final_noise_pred}
%\mathbf{\Bar{\epsilon}}_\theta(&\mathbf{z}_t, \mathbf{c}_p, \mathbf{c}_S)= \nonumber \\
       &\mathbf{\epsilon}_\theta(\mathbf{z}_t) + s_g \big(\mathbf{\epsilon}_\theta(\mathbf{z}_t, \mathbf{c}_p) - \mathbf{\epsilon}_\theta(\mathbf{z}_t)- \gamma(\mathbf{z}_t, \mathbf{c}_p, \mathbf{c}_S)\big) 
\end{align}
By doing so, SLD could reduce the possibility of generating content described by the inappropriate concept. While SLD could only reduce the generation in a certain direction, we could further upgrade the method to a \textit{multi-directional} version by combining different text descriptions defined by target attributes with different weights, which is further used by our model to guide the generation of bias-free samples.

% \paragraph{Background.} Our problem can be viewed as a black-box optimization task. \textit{We have attempted to apply classical frameworks, such as the distributed adaptive signal fusion (DASF) algorithm, the iterative proportional fitting (IPF) method, the simulated moment method, and the fixed-point method, but none of them worked well due to the high non-convexity and non-linearity of the diffusion mapping}. Therefore, we resort to the classical black-box reinforcement framework, which is commonly used for online hyperparameter optimization problems. We can also regard the diffusion-debiasing process as an online hyperparameter optimization problem.

\paragraph{Multi-directional Guidance.} 
We use weights $\bm{a} = \{{a}_0, ..., {a}_{n-1}\}$ as coefficients for text latent vectors $\{{g}_0, ..., {g}_{n-1}\}$ to obtain an unsafe vector as the multi-directional guidance: 
\begin{equation}
\bm{u} = \sum_{i=0}^{n} {a}_i {g}_i,
\end{equation}
which replaces the inappropriate concept vector $\mathbf{c}_S$ in SLD. Our goal is to optimize $\bm{a}$ to achieve the balance between various attributes and generate bias-free samples.
% consider a diffusion model that generates images from text prompts. The model takes as input a noise latent vector $\bm{z}$, a text prompt $\bm{p}$, and a set of class latent vectors $\{\bm{x}_0, ..., \bm{x}_n\}$ with corresponding weights $\{[\bm{a}]_0, ..., [\bm{a}]_n\}$. The class latent vectors represent different attributes of the images, such as gender and skin color. The model combines the class latent vectors and weights to produce an unsafe latent vector, which influences the diffusion process. The objective is to make the output images have a balanced distribution of attributes, i.e., as close to uniform as possible.

\paragraph{Weight Optimization.}
To simplify the understanding of the diffusion mapping process from input variables $\{\bm{z}, \bm{p}, \bm{u}\}$ to the final output, we can treat the processes of softmax-normalization for $\bm{a}$, unsafe latent vector generation, and guided diffusion as a black-box mapping. This mapping takes the input weight vector $\bm{a} = \{{a}_0, ..., {a}_{n-1}\}$ and produces the class-wise frequency statistics $\bm{s} \in \mathbb{R}_+^n$. Mathematically, we can represent this as: 
\begin{equation} \label{eq7}
\bm{s} = \text{Black-Box}(\bm{a}),
\end{equation}
where ${a}_i$ represents the $i$-th element. 
To evaluate the performance of this mapping, we define a loss function $E$ using the KL divergence between the normalized frequency vector $\bm{s}$ and a uniform distribution as follows: 
\begin{equation} \label{eq8}
E(\bm{s}) = D_{KL}(\Bar{\bm{s}}, U(n))=\sum_i^n \Bar{\bm{s}}_i \log (n*\Bar{\bm{s}}_i),
\end{equation}
where $\Bar{\bm{s}}$ is the normalized frequency vector of $\bm{s}$,  $\Bar{\bm{s}}_i = \frac{\bm{s}_i}{\sum_j \bm{s}_j}$, and $U(n)$ is a uniform distribution with n values of equal probability 1/n.
The goal is to find the optimal weight vector $\bm{a}^*$ that minimizes the loss function: 
\begin{equation}
\bm{a}^* = \arg \min_{\bm{a}} E \Big(\text{Black-Box}(\bm{a}) \Big).
\end{equation}

  \renewcommand{\algorithmicrequire}{ \textbf{Input:}} 
  \renewcommand{\algorithmicensure}{ \textbf{Output:}}

\subsection{Debiasing via Weight Optimization}
Our goal is to obtain a set of weights $\bm{a}$ that could debias the output samples from diffusion models and achieve a balanced distribution. In Sec. \ref{Reinforcement Solver}, we introduce a naive policy gradient reinforcement solver, which we found to be sub-optimal in our experiments. Further, in Sec. \ref{IDA}, we propose an iterative distribution alignment method, which turns out to successfully resolve the bias in diffusion models.

\subsubsection{Reinforcement Solver} \label{Reinforcement Solver}

% \thr{Ruifei, please see the comments here.} 
To optimize our model, we employ the reinforcement solver approach. We instantiate a distribution $\pi_\mathbf{A}$ with parameters $\mathbf{A} \in \mathbb{R}^n$ and sample $\bm{a}$ from this distribution, denoted as $\bm{a} \sim \pi_\mathbf{A}$. The distribution $\pi_\mathbf{A}$ is formed using a Gaussian function centered at $\mathbf{A}$. Treating all variables independently, we can express $\pi_\mathbf{A}(a)$ as: 
\begin{equation}
\pi_\mathbf{A}(a) = \prod_i^n G([\mathbf{A}]_i,1),
\end{equation}
where $G([\mathbf{A}]_i,1)$ represents a Gaussian distribution with a mean of $[\mathbf{A}]_i$ and a standard deviation of 1. The elements of $\mathbf{A}$ are randomly initialized.

The algorithm is shown in Algorithm \ref{algo:rs}. 
For each iteration indexed by $t$:
We first draw $K$ candidates $\bm{a}_0,...\bm{a}_{K-1}$ from $\pi_{\mathbf{A}_t}$. Then for each candidate $\bm{a}_k$, we feed it into the black box and obtain the reward $R_{t, k} = exp(-E_{t, k})$. We collect the loss and reward for all candidates and the stopping criterion of the algorithm is when the collected loss is below a pre-defined threshold T. Afterwards, we compute the policy gradient and update the parameters as described in Algorithm \ref{algo:rs} (Line \ref{a1:line8} -- \ref{a1:line9}). 

To improve the stability and reduce the variance of our pipeline, we can apply the following techniques:

1. Use a momentum term.

2. Ensure that $R_{t, 1}, ...R_{t, k}$ have a zero mean. By doing so, we obtain the modified update rule: 
\begin{equation}
\bm{A}_t = \bm{A}_t + \eta \sum_{k=1}^K (R_{t, k} - v_t) \cdot \frac{\partial \pi_{\bm{A}}(\bm{a}_k)}{\partial \bm{A}},
\end{equation}
where $v_t$ indicates the min or the mean reward at the $t$-th step. 

In our experiments, the reinforcement solver approach shows slow convergence speed and unfavorable effectiveness in resolving the bias, and thus we resort to designing more intuitive methods.

  \renewcommand{\algorithmicrequire}{ \textbf{Input:}} 
  \renewcommand{\algorithmicensure}{ \textbf{Output:}}

\begin{algorithm} 
    \footnotesize
    \caption{Reinforcement Solver}
    \label{algo:rs}
    \begin{algorithmic}[1]
        \Require
        An initial distribution $\pi_\mathbf{A}$ with parameters $\mathbf{A}$, the Black-Box model Black-Box(), total iterations N, number of candidates K, update parameter $\eta$, loss threshold T.
        \Ensure
        Updated parameters $\mathbf{A}$. 

\State \textbf{for} {$t=0,1,\ldots,N-1$} \textbf{do} 
\State \hspace{0.4cm} Draw $K$ candidates from $\bm{a}_0,...\bm{a}_{K-1} \sim \pi_{\mathbf{A}_t}$.
\State \hspace{0.4cm} \textbf{for} {$k=0,1,\ldots,K-1$} \textbf{do} 
\State \hspace{0.8cm} \textcolor{blue}{ \# According to Eq. \ref{eq7} and Eq. \ref{eq8} }
\State \hspace{0.8cm} Feed $\bm{a}_k$ into the black box and obtain KL loss $E_{t, k}$. 
\State \hspace{0.8cm} The reward is $R_{t, k} = exp(-E_{t, k})$
\State \hspace{0.4cm} \textbf{end for}
\State \hspace{0.4cm} Sum up: $\mathcal{E}(R_t) = \sum_k R_{k} \pi_{\bm{A}}(\bm{a}_k)$, $\mathcal{E}(E_t) = \sum_k E_{t, k}$
\State \hspace{0.4cm}\textcolor{blue}{ \# Stopping Criterion}
\State \hspace{0.4cm} \textbf{If} $\mathcal{E}(E_t) < T$: \textbf{return} $\bm{A}_t$
\State \hspace{0.4cm}\textcolor{blue}{ \# Calculate the policy gradient}
\State \hspace{0.4cm} $\nabla_{\bm{A}} \mathcal{E} (R_t) = \sum_{k=1}^K R_{t, k} \cdot \frac{\partial \pi_{\bm{A}}(\bm{a}_k)}{\partial \bm{A}}$. \label{a1:line8}
\State \hspace{0.4cm}\textcolor{blue}{ \# Update the parameters by the policy gradient}
\State \hspace{0.4cm}
$\bm{A}_t = \bm{A}_t + \eta \nabla_{\bm{A}} R_t = \bm{A}_t + \eta \sum_{k=1}^K R_{t, k} \cdot \frac{\partial \pi_{\bm{A}}(\bm{a}_k)}{\partial \bm{A}}$. \label{a1:line9}
\State \textbf{end for}
    \end{algorithmic}
    \label{alg:distribution}
\end{algorithm}

   \begin{figure*} 
% \vspace{-1.3cm}
   \begin{center}
   % \fbox{\rule{0pt}{2in} \rule{.9\linewidth}{0pt}} 
   \includegraphics[width=.9\linewidth]{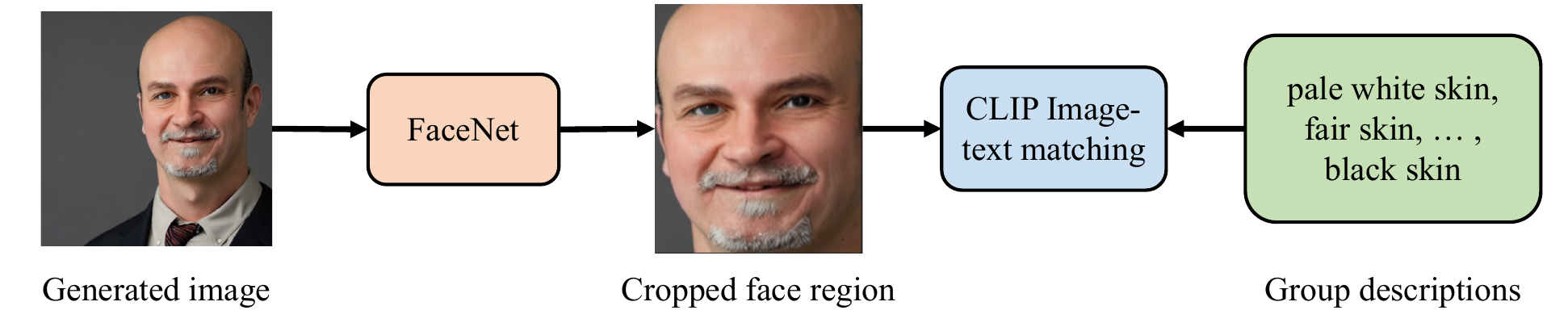} 
   \end{center}
    % \vspace{-0.4cm}
      \caption{\textbf{Classifier}. We use a face detector (FaceNet) to crop the face region of generated images, and then match them with group descriptions via CLIP image-text similarity.}
   \label{fig: classifier} 
   % \vspace{-0.5cm} 
   \end{figure*}

\subsubsection{Iterative Distribution Alignment}
\label{IDA}
\begin{algorithm} 
    \footnotesize
    \caption{Iterative Distribution Alignment}
    \label{algo:ida}
    \begin{algorithmic}[1]
        \Require
        The Black-Box model Black-Box(), total iterations N, update parameter $\alpha$, number of attributes n, loss threshold T.
        \Ensure
        Desired $\bm{a}$. 
\State At $t=0$, set all $a_i^{(0)}=0$. Feed $\bm{a}^{(0)}$ into the black box and obtain the frequency $\bm{s^{(0)}}$.
\State At $t=1$, set ${a}_i^{(1)}=\Bar{{s}_{i}}^{(0)} - 1/n$. Feed $\bm{a}^{(1)}$ into the black box and obtain the frequency $\bm{s^{(1)}}$.
\State \textbf{for} {$t=2,\ldots,N-1$} \textbf{do} 
\State \hspace{0.4cm}\textcolor{blue}{ \# Residual Learning}
\State \hspace{0.4cm} We set ${a}_i^{(t)}={a}_i^{(t-1)}+\alpha * (\Bar{{s}_{i}}^{(t-1)} - 1/n)$.
\State \hspace{0.4cm} We feed $\bm{a}^{(t)}$ into the black box and obtain the frequency $\bm{s^{(t)}}$ as well as the KL loss $E_{t}$.
\State \hspace{0.4cm}\textcolor{blue}{ \# Stopping Criterion}
\State \hspace{0.4cm} \textbf{If} $(E_t) < T$: \textbf{return} $\bm{a}^{(t)}$. \label{a2:line9}
\State \textbf{end for}
    \end{algorithmic}
    \label{alg:distribution}
\end{algorithm}
   
In this part, we design an iterative distribution alignment method using the residual from each $\Bar{\bm{s}}_i$ to the uniform value $1/n$. As shown in Figure \ref{fig: framework}, we iteratively use the frequency obtained from generated images to feedback to the weights by residual learning. 

Concretely, we present our algorithm in Algorithm \ref{algo:ida}. In each iteration except the first one, we use the residual from the frequency of the last iteration to the uniform value as a update clue. Intuitively, this punishes the attributes that have above-average frequency since they are recognized as unsafe directions, while below-average attributes would be encouraged.

% $\quad$(1). At iteration $t$=0, we set all $\bm{a}_i^{(0)}=0$ and generate images to obtain the frequency $\bm{s^{(0)}}$ from a classifier module (explained in the following).

% $\quad$(2). At iteration $t$=1, we set $\bm{a}_i^{(1)}=\Bar{\bm{s}_{i}}^{(0)} - 1/n$. Intuitively, this punishes the attributes that have above-average frequency since they are recognized as unsafe directions, while below-average attributes would be encouraged. With $\bm{a}_i^{(1)}$, we forward to the black box to obtain $\bm{s^{(1)}}$.

% $\quad$(3). At iteration $t$=k, we set $\bm{a}_i^{(k)}=\bm{a}_i^{(k-1)}+\alpha * (\Bar{\bm{s}_{i}}^{(k-1)} - 1/n)$, where $\alpha$ is a hyper-parameter. We iteratively learn the weights from the residual.

% $\quad$(4). We iterate this process until the KL divergence loss $E(\bm{s})$ is desired.

Despite its simplicity, the proposed method works surprisingly well in our experiments. With only 1-3 iterations, the KL divergence could be largely reduced and achieve favorable performance.

\paragraph{Classifier} is an automatic procedure to classify the generated images into different groups. As shown in Figure \ref{fig: classifier}, we first use a face detector to crop out the face region of an image to exclude the interference of background for better classification. Next, we use the CLIP \citep{radford2021learning} model to compute the cosine similarity between cropped images with text descriptions of different groups. 

\begin{figure*} 
% \vspace{-0.95cm}
   \begin{center}
   % \fbox{\rule{0pt}{2in} \rule{.9\linewidth}{0pt}} 
   \includegraphics[width=1.0\linewidth]{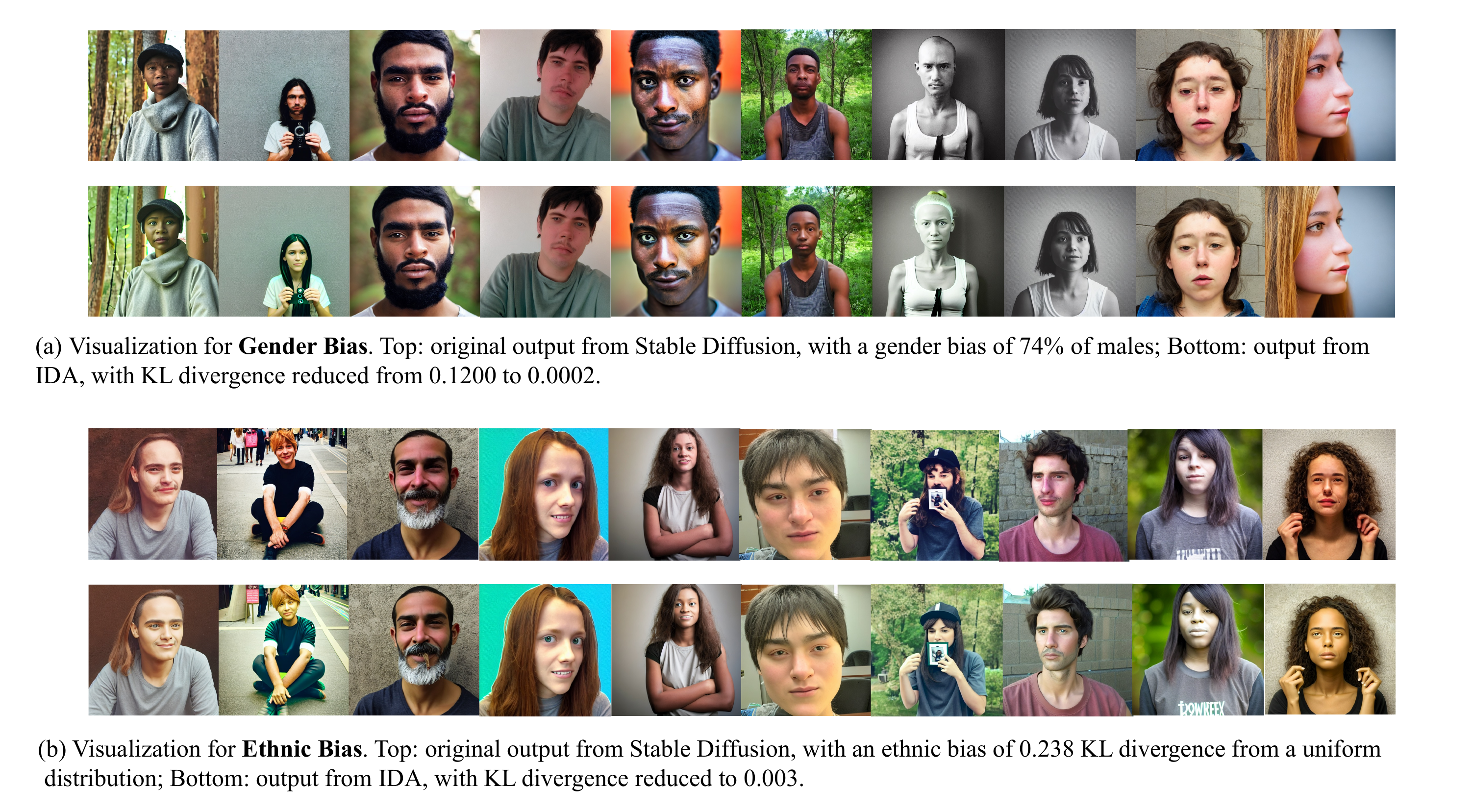} 
   \end{center}
    % \vspace{-0.4cm}
      \caption{\textbf{Visualization of IDA Debiasing.} IDA successfully redistributes the gender and ethnic distribution to a balanced distribution.}
   \label{fig: vis} 
   % \vspace{-0.5cm} 
   \end{figure*}

\section{Experiments}
\subsection{Experimental Settings} 
In this work, we evaluate two settings for social bias: gender bias and ethnic bias. For gender bias, we only consider two genders, male and female. For ethnic bias, we adopt the Fitzpatrick Skin Scale classification system \citep{fitzpatrick1988validity}, which divides people into six groups by skin type.  
\paragraph{Implementation Details.}
For the diffusion model, we experiment with the state-of-the-art Stable Diffusion \citep{rombach2022high}. For each iteration, we generate 100 images using prompts including "person" but without any preference about gender and ethnicity. All the implementations of the hyper-parameters in the safe latent diffusion algorithm are set to the same as the "Hyp-Weak" variant as in its paper \citep{schramowski2023safe} except the guidance scale is reduced to 1$\sim$2. To obtain the frequency from the images, we first use FaceNet \citep{schroff2015facenet} to crop the face region of the generated images, and then use CLIP \citep{radford2021learning} to match the cropped images with text of different attributes.

\subsection{Main Results.}
\paragraph{Gender Bias.}
In our experiments, without any debiasing, the original Stable Diffusion exhibits a bias towards masculinity, with a frequency of 74\% of males and only 26\% of females. Fortunately, as shown in Table \ref{tab: gender ida}, with the proposed iterative distribution alignment (IDA) method, after iteration $t=1$, the frequency redistributes to 48\% of males and  52\% of females, and the KL divergence significantly reduces from 0.12 to 0.0008. And one more iteration leads to better debiasing results. We found the proposed method works fast and well for the gender bias.

\renewcommand\arraystretch{1.2}
  \begin{table}[htbp] 
   \centering
   % \vspace{-0.2cm}
   \begin{small}
       \setlength\tabcolsep{18pt} 
       \begin{tabular}{cccc}
           \bottomrule[1pt]
           & Male & Female & KL  \\ 
           \hline
           t=0 & 0.74& 0.26&  0.1200   \\
           t=1 & 0.48& 0.52 &  0.0008 \\
           t=2 & 0.49 & 0.51 & \highlight{0.0002} \\
           \toprule[0.8pt] 
       \end{tabular}
   \end{small}
   % \vspace{-0.3cm}  
   \caption{\textbf{Gender Bias.} Frequency of genders at different iterations in IDA. KL: KL divergence with a uniform distribution.}
   % \vspace{-0.3cm}
   \label{tab: gender ida}
\end{table}

\paragraph{Ethnic Bias.} The Fitzpatrick Skin Scale \citep{fitzpatrick1988validity} divides people into six skin types: (\uppercase\expandafter{\romannumeral1}) pale white skin, (\uppercase\expandafter{\romannumeral2}) fair skin, (\uppercase\expandafter{\romannumeral3}) olive skin, (\uppercase\expandafter{\romannumeral4}) moderate brown skin, (\uppercase\expandafter{\romannumeral5}) dark brown skin, and (\uppercase\expandafter{\romannumeral6}) black skin. 
As shown in Table \ref{tab: ethnic ida}, the original distribution ($t=0$) has a bias towards whiteness (group \uppercase\expandafter{\romannumeral1} and \uppercase\expandafter{\romannumeral2} have a relatively major population). Effectively, with only one iteration, the proposed IDA method largely reduces the KL divergence from 0.238 to 0.007, making the biased distribution shift to a uniform one. And with another iteration, IDA further improves the KL divergence to 0.003. 

\renewcommand\arraystretch{1.2}
  \begin{table}[htbp] 
   \centering
   % \vspace{-0.2cm}
   \begin{small}
       \setlength\tabcolsep{5pt} 
       \begin{tabular}{l cccccc c}
           \bottomrule[1pt]
           & \uppercase\expandafter{\romannumeral1} & \uppercase\expandafter{\romannumeral2} & \uppercase\expandafter{\romannumeral3} & \uppercase\expandafter{\romannumeral4}  & \uppercase\expandafter{\romannumeral5} & \uppercase\expandafter{\romannumeral6} & KL  \\ 
           \hline
           t=0 & 0.384& 0.242& 0.061& 0.040&0.141&0.131 & 0.238   \\
           t=1 & 0.172& 0.182 & 0.172& 0.192 & 0.141& 0.141& 0.007 \\
           t=2 & 0.192 & 0.172 & 0.162 & 0.172& 0.152 & 0.152 & \highlight{0.003} \\
           \toprule[0.8pt] 
       \end{tabular}
   \end{small}
   % \vspace{-0.3cm}  
   \caption{\textbf{Ethnicity Bias.} Frequency of six skin types at different iterations in IDA. KL: KL divergence with a uniform distribution.}
   % \vspace{-0.3cm}
   \label{tab: ethnic ida}
\end{table}

\begin{figure*} 
% \vspace{-0.95cm}
   \begin{center}
   % \fbox{\rule{0pt}{2in} \rule{.9\linewidth}{0pt}} 
   \includegraphics[width=1.0\linewidth]{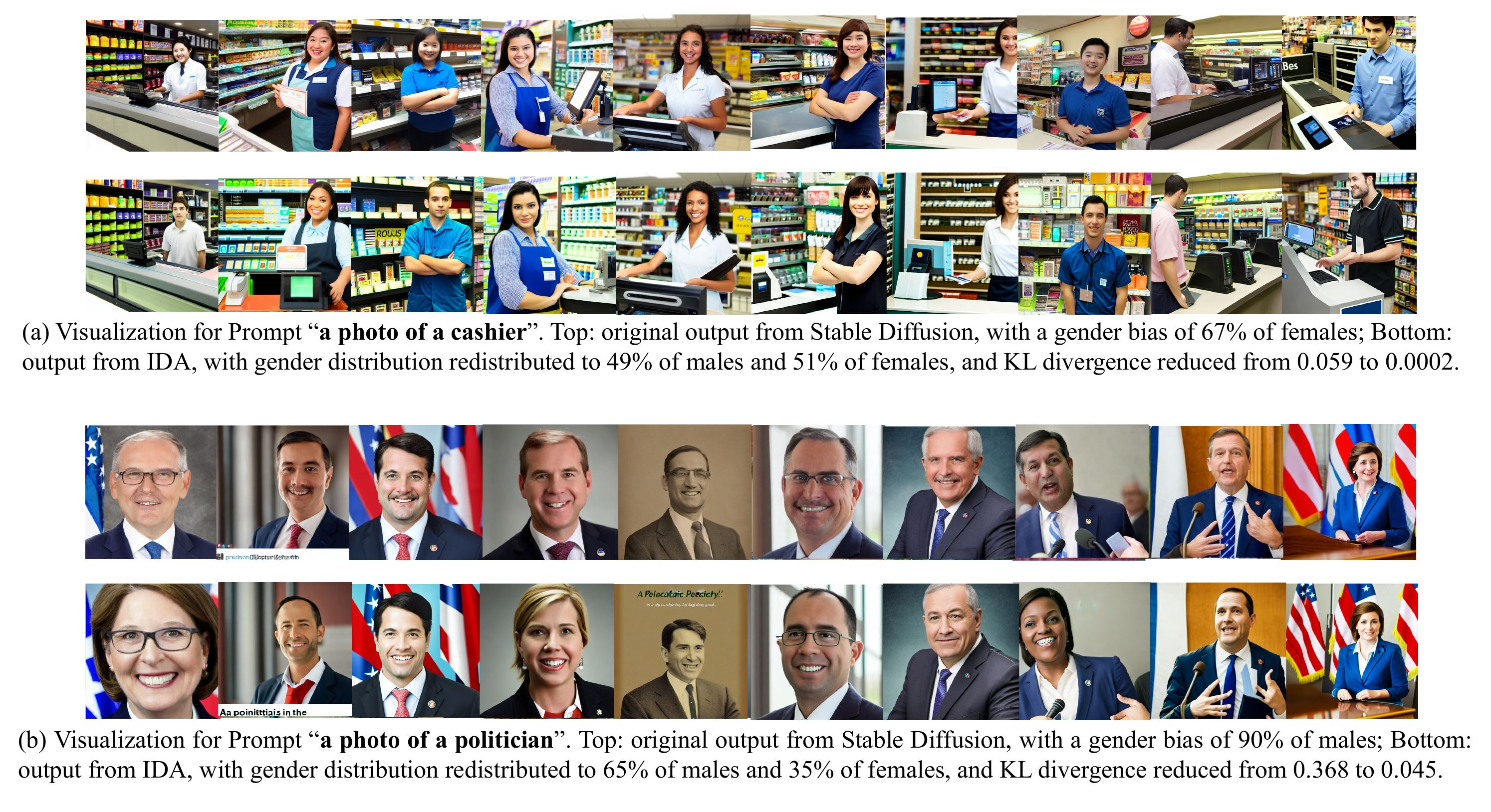} 
   \end{center}
    % \vspace{-0.4cm}
      \caption{\textbf{Gender Bias Study on Occupations.} IDA successfully reduce the gender bias for different occupations.}
   \label{fig: occu} 
   % \vspace{-0.5cm} 
   \end{figure*}

\paragraph{Visualization of IDA Debiasing.}
In Figure \ref{fig: vis}, we demonstrate the visual effects of how the proposed IDA method could help to resolve the gender and ethnic bias. In Figure \ref{fig: vis} (a), we can see that IDA could turn some images of males into females while maintaining the major structure of the image, and generally control the distribution of generated images to be balanced between males and females. In Figure \ref{fig: vis} (b), we could observe that IDA changes the skin color of the person in the generated images and again maintains the major structure of the image. In conclusion, IDA could help redistribute the distribution to a balanced one while not affecting the structure of the generated images.

\begin{figure} 
% \vspace{-0.95cm}
   \begin{center}
   % \fbox{\rule{0pt}{2in} \rule{.9\linewidth}{0pt}} 
   \includegraphics[width=1.0\linewidth]{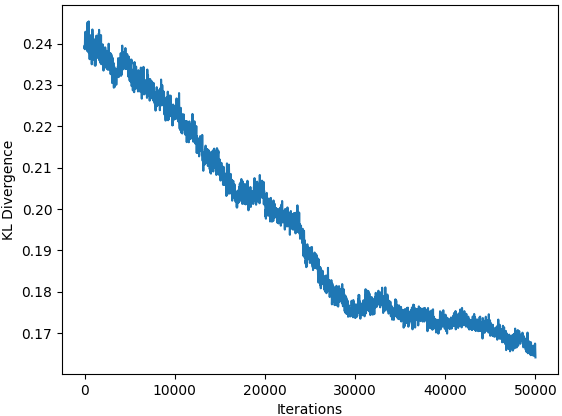} 
   \end{center}
    % \vspace{-0.4cm}
      \caption{\textbf{Reinforcement Solver}: KL v.s no. iterations. The algorithm takes many iterations to converge and the KL divergence only reduces from around 0.24 to 0.16, the improvement of bias is limited.}
   \label{fig: RL} 
   % \vspace{-0.5cm} 
   \end{figure}

\paragraph{Effectiveness of Reinforcement Solver.}
We first experiment with different hyper-parameters of population $K$ = \{5, 10, 20, 40, 60\}, and iterations of \{3, 10, 50\}, but all the results remain similar with the original poor biased distribution. Since the running cost grows linearly with the population $K$ and the number of iterations, it becomes hard to directly conduct experiments with larger $K$ and iterations (1 iteration for $K=1$ takes around 5.4 minutes on an A100 GPU, thus 50 iterations with $K=60$ would take over 11 days on an A100 GPU.) 
For this, we conduct a simulation experiment, where we use a simple MLP with Softmax to simulate the frequency of the attribute distribution. As shown in Figure \ref{fig: RL}, with a population of $K=60$, the Reinforcement Solver needs over 50k iterations to converge, and the KL divergence has not been reduced to a relatively low level, which shows the slow convergence and inefficiency of the algorithm. Such a large number of iterations would be impractical in realistic situations. In contrast, the proposed IDA method only needs 1$\sim$3 iterations to achieve a surprisingly small KL divergence, only taking around 5$\sim$20 minutes on an A100 GPU.

\renewcommand\arraystretch{1.2}

\paragraph{Image Fidelity and Text Alignment.} Here, we conduct experiments to evaluate whether our method would affect generated image quality and text alignment. As shown in Table \ref{tab: fid}, we follow previous research \citep{schramowski2023safe} and report the COCO FID-30k scores and CLIP distance. Our method does not affect image fidelity and text alignment.

% \begin{table}
%     \centering
%     \small
%     \setlength\tabcolsep{3pt} 
%     \begin{tabular}{l  c c}
%     \textbf{Model} & \textbf{Image Fidelity} (FID-30k $\downarrow$) & \textbf{Text Alignment} (CLIP $\downarrow$) \\
%       \hline
%     SD    & 14.43 & 0.75 \\
%     Ours & 15.21 & 0.75 \\
%     \end{tabular}
%     \caption{\textbf{Image fidelity and text alignment.} Our method does not influence image fidelity and text alignment.}
%     \label{tab: fid}
% \end{table}

\renewcommand\arraystretch{1.2}
  \begin{table}[htbp] 
   \centering
   % \vspace{-0.2cm}
   \begin{small}
       \setlength\tabcolsep{16pt} 
       \begin{tabular}{lcc}
           \bottomrule[1pt]
           \multirow{ 2}{*}{\textbf{Model}} & \textbf{Image Fidelity}  & \textbf{Text Alignment} \\ 
           & FID-30k $\downarrow$ & CLIP $\downarrow$ \\
           \hline
           SD    & 14.43 & 0.75 \\
            Ours & 15.21 & 0.75 \\
           \toprule[0.8pt] 
       \end{tabular}
   \end{small}
   % \vspace{-0.3cm}  
   \caption{\textbf{Image fidelity and text alignment.} Our method does not influence image fidelity and text alignment.}
   % \vspace{-0.3cm}
   \label{tab: fid}
\end{table}

\subsection{Gender Bias Study on Occupations}
Our main experiments use a prompt with the main subject of "person" but without specifying the occupation of the person. However, in Figure \ref{fig: bias eg}, we observe that different occupations have different stereotypes about gender, resulting in different gender biases for different occupations (e.g. males seem to taking the lead in high-paying occupations like CEO, politician, and females seem to be the majority in low-paying occupations such as cashier, housekeeper). 

Hence, in this part, we dive into the gender bias of different occupations. Concretely, we experiment with six occupations, three relatively high-paying occupations (CEO, politician, professor) and three relatively low-paying occupations (cashier, housekeeper, teacher). The results are shown in Table \ref{tab: occupation}. Note that gender bias could sometimes be very extreme --- one gender may take up all the percentage for a certain occupation. For both high-paying and low-paying occupations, the proposed IDA method successfully reduces gender bias and significantly decreases the KL divergence between the gender distribution with a uniform distribution.  

Further, we provide visualization examples of occupations of cashier and politician in Figure \ref{fig: occu}. In Figure \ref{fig: occu} (a) Top, we could see that around 7 out of 10 images of cashiers are females, and with our IDA method, we redistribute the gender distribution to a nearly uniform distribution, turning 20\% female cashiers into male cashiers (Figure \ref{fig: occu} (a) Bottom). For the politician case, the gender bias from Stable Diffusion is more severe, with males taking up to 90\%. As shown in Figure \ref{fig: occu} (b), our IDA method alleviates the extreme biased distribution and turns around 25\% of the male figures into female ones.

\renewcommand\arraystretch{1.2}
  \begin{table}[htbp] 
   \centering
   % \vspace{-0.2cm}
   \begin{small}
       \setlength\tabcolsep{10pt} 
       \begin{tabular}{ccccc}
           \bottomrule[1pt]
           Occupation& Model& Male & Female & KL  \\ 
           \hline
           \multirow{ 2}{*}{CEO} & SD & 1.00 & 0.00 &  0.6931   \\
           & IDA & 0.64& 0.36 &  \highlight{0.0328} \\
           \hline
           \multirow{ 2}{*}{Politician} & SD & 0.90 & 0.10 &  0.3680   \\
           & IDA & 0.65& 0.35 &  \highlight{0.0457} \\
           \hline
           \multirow{ 2}{*}{Professor} & SD &1.00 & 0.00 &  0.6931   \\
           & IDA & 0.57& 0.43 &  \highlight{0.0098} \\
           \hline
           \hline
           \multirow{ 2}{*}{Cashier} & SD &0.33 & 0.67 &  0.0590   \\
           & IDA & 0.49& 0.51 &  \highlight{0.0002} \\
           \hline
           \multirow{ 2}{*}{Housekeeper} & SD &0.02 & 0.98 &  0.5951   \\
           & IDA & 0.46& 0.54 &  \highlight{0.0032} \\
           \hline
           \multirow{ 2}{*}{Teacher} & SD &0.20 & 0.80 &  0.1927   \\
           & IDA & 0.45& 0.55 &  \highlight{0.0050} \\
           \toprule[0.8pt] 
       \end{tabular}
   \end{small}
   % \vspace{-0.3cm}  
   \caption{\textbf{Gender Bias of different occupations.} Frequency of genders of the generated images from Stable Diffusion (SD) or our Iterative Distribution Alignment (IDA) method. KL: KL divergence with a uniform distribution.}
   % \vspace{-0.3cm}
   \label{tab: occupation}
\end{table}

\subsection{Discussion and Limitations}
\paragraph{Efficiency.} Compared with the Reinforcement Solver method, our IDA method only needs 1$\sim$3 iterations to achieve convergence and usually only a single iteration could achieve promising results. However, we claim that our method manages to resolve the bias for each given prompt, while once the prompt changes, the algorithm needs to be re-run, which may be costly in practice.

\paragraph{Explainability.} Though the proposed IDA method achieves fast convergence and promising results, it lacks formal proof or explanation of its success. We infer the reason to be some characteristics of Stable Diffusion, maybe related to its potential positive correlation property between its output and conditional text embedding due to its training process linearly combining unconditioned prediction and conditioned prediction.

\section{Conclusion}
In this work, we systematically investigate the problem of resolving the social bias in text-to-image diffusion models from gender and ethnicity perspectives. Our experimental results demonstrate our proposed IDA method could effectively resolve the bias within practical GPU hours (up to 20 minutes on an A100). We also point out our limitations on efficiency and explainability which we regard as future work.

\paragraph{Broader Impact.} We emphasize the importance of addressing the social bias exhibited in text-to-image diffusion models, which is under-explored. We hope our work could inspire more future works on aligning text-to-image generative AI with social ethics.

\nocite{langley00}

\bibliography{example_paper}
\bibliographystyle{icml2024}

%%%%%%%%%%%%%%%%%%%%%%%%%%%%%%%%%%%%%%%%%%%%%%%%%%%%%%%%%%%%%%%%%%%%%%%%%%%%%%%
%%%%%%%%%%%%%%%%%%%%%%%%%%%%%%%%%%%%%%%%%%%%%%%%%%%%%%%%%%%%%%%%%%%%%%%%%%%%%%%
% APPENDIX
%%%%%%%%%%%%%%%%%%%%%%%%%%%%%%%%%%%%%%%%%%%%%%%%%%%%%%%%%%%%%%%%%%%%%%%%%%%%%%%
%%%%%%%%%%%%%%%%%%%%%%%%%%%%%%%%%%%%%%%%%%%%%%%%%%%%%%%%%%%%%%%%%%%%%%%%%%%%%%%
% \newpage
% \appendix
% \onecolumn

% You can have as much text here as you want. The main body must be at most $8$ pages long.
% For the final version, one more page can be added.
% If you want, you can use an appendix like this one, even using the one-column format.
%%%%%%%%%%%%%%%%%%%%%%%%%%%%%%%%%%%%%%%%%%%%%%%%%%%%%%%%%%%%%%%%%%%%%%%%%%%%%%%
%%%%%%%%%%%%%%%%%%%%%%%%%%%%%%%%%%%%%%%%%%%%%%%%%%%%%%%%%%%%%%%%%%%%%%%%%%%%%%%

\end{document}